\documentclass{article}

\usepackage[final,nonatbib]{neurips_bdl2019}
\usepackage[utf8]{inputenc} 
\usepackage[T1]{fontenc}    
\usepackage{hyperref}       
\usepackage{url}            
\usepackage{booktabs}       
\usepackage{amsfonts}       
\usepackage{nicefrac}       
\usepackage{microtype}      
\usepackage{amsmath}
\usepackage{amssymb}
\usepackage{mathtools} 
\usepackage{tikz}
\usepackage[numbers]{natbib}
\usepackage{blindtext}
\usepackage{mathtools}
\usepackage{relsize}

\usetikzlibrary{decorations.pathreplacing}
\graphicspath{{./figures/}}

\author{Joshua C. Chang \\ Epidemiology and Biostatistics \\ Rehabilitation Medicine, CC, NIH \\ 
\&  team {med$\varepsilon_{\text{rrata}}$} \\
\href{mailto:josh@mederrata.com}{josh@mederrata.com} 
\\ \and Shashaank Vattikuti  \& Carson C. Chow \\Laboratory of Biological Modeling, NIDDK, NIH \\
 \&  team {med$\varepsilon_{\text{rrata}}$} \\
\texttt{\{\href{mailto:	shashaank.vattikuti@nih.gov}{shashaank.vattikuti}, \href{mailto:carsonc@niddk.nih.gov}{carson.chow}\}}@nih.gov \\}

\title{Probabilistically-autoencoded horseshoe-disentangled multidomain item-response theory models}

\begin{document}

\maketitle

\begin{abstract}
Item response theory (IRT) is a non-linear generative probabilistic paradigm for using exams  to identify, quantify, and compare latent traits of individuals, relative to their peers, within a population of interest. 
In pre-existing multidimensional IRT methods, one requires a factorization of the test items.
 For this task, linear exploratory factor analysis is  used, making IRT a posthoc model.
 We propose skipping the initial factor analysis by using a sparsity-promoting horseshoe prior to perform factorization directly within the IRT model so that all training occurs in a single self-consistent step. Being a hierarchical Bayesian model, we adapt the WAIC to the problem of dimensionality selection. IRT models are analogous to probabilistic autoencoders. 
 By binding the generative IRT model to a Bayesian neural network (forming a probabilistic autoencoder), one obtains a scoring algorithm consistent with the interpretable Bayesian model. In some IRT applications the black-box nature of a neural network scoring machine is desirable.  
 In this manuscript, we demonstrate within-IRT factorization and comment on scoring approaches.
\end{abstract}

\section{Introduction}

\subsection{Item response theory (IRT)}

IRT~\cite{chang_item_2005,fries_item_2014} is the dominant statistical paradigm for using tests in order to quantify latent traits for individuals in the context of their populations. In both form and function, IRT models resemble autoencoders.  Like in an autoencoder, the traits are a nonlinear factor representation of input data, and the latent representation is a major analytical product of interest. While additional latent variables exist both in IRT and in neural network-based autoencoders, of these, only in an IRT model are the latent variables interpretable. These parameters constitute item calibrations -- difficulty and discrimination parameters -- that correspond to the information content of each test question. In effect, IRT models are able to separate individual effects from test-specific effects in understanding responses. In contrast, the weights of a neural network lack mechanistic interpretability.

Model interpretability is a weakness for deep learning because artificial neural networks are intrinsically black boxes. It is difficult to understand how neural networks encode relationships between variables in their multidimensional mappings. In practical terms, this opaqueness has made it difficult to determine whether neural networks are making good objective inferences, or are simply codifying biases inherent in their training data.

Unfortunately, these issues are an impediment to using autoencoders in testing, where testing outcomes have real-life consequences.
Disentangled autoencoders like the $\beta$-variational autoencoder~\cite{higgins_beta-vae:_2016} ($\beta$-VAE), and related methods~\cite{kim_disentangling_2018,antoran_disentangling_2019,li_disentangled_2018,burgess_understanding_2018} show promise -- they provide improved interpretability by producing disentangled representations. What is meant by disentanglement, however, is specific to the method.  Disentanglement typically refers to statistical independence of these latent factors, and not on dependencies to the input data itself.

Typically, for IRT, one desires disentanglement of original test items through partitioning, so that each partition informs a different dimension of the latent trait. The partitioning is  performed prior to training of the IRT model, using exploratory factor analysis.  Subdomains of the IRT model are then fit independently, conditional on the partition, using various empirical Bayesian~\cite{bock_irt_1997} or  Bayesian methods~\cite{weng_real-time_2018,vispoel_applications_2018}. 

Hence, IRT models are dependent on the initial partitioning of items into domains.
The domain partitioning is a linear covariance analysis, where various criteria are commonly-used for accepting and rejecting items within each factor domain, without attention to consistency with the nonlinear IRT model. Additionally, partitional uncertainty is not propagated into the IRT model.

We propose factorization directly within IRT models using sparse coding. Additionally, we interpret IRT as a probabilistic decoder, and couple it to a neural network encoder.  In training the encoder, the neural network learns the inversion of the generative model. Hence, one obtains a scoring machine that is consistent with the learned posterior of the IRT model. Testing is a rare application in which black box scoring machines may have value. The uninterpretable nature of the neural network can help impede gaming of exams if parameters from the scoring algorithm become partially compromised.

\section{Modeling}

\begin{figure}[h]
    \centering
\includegraphics[width=\linewidth]{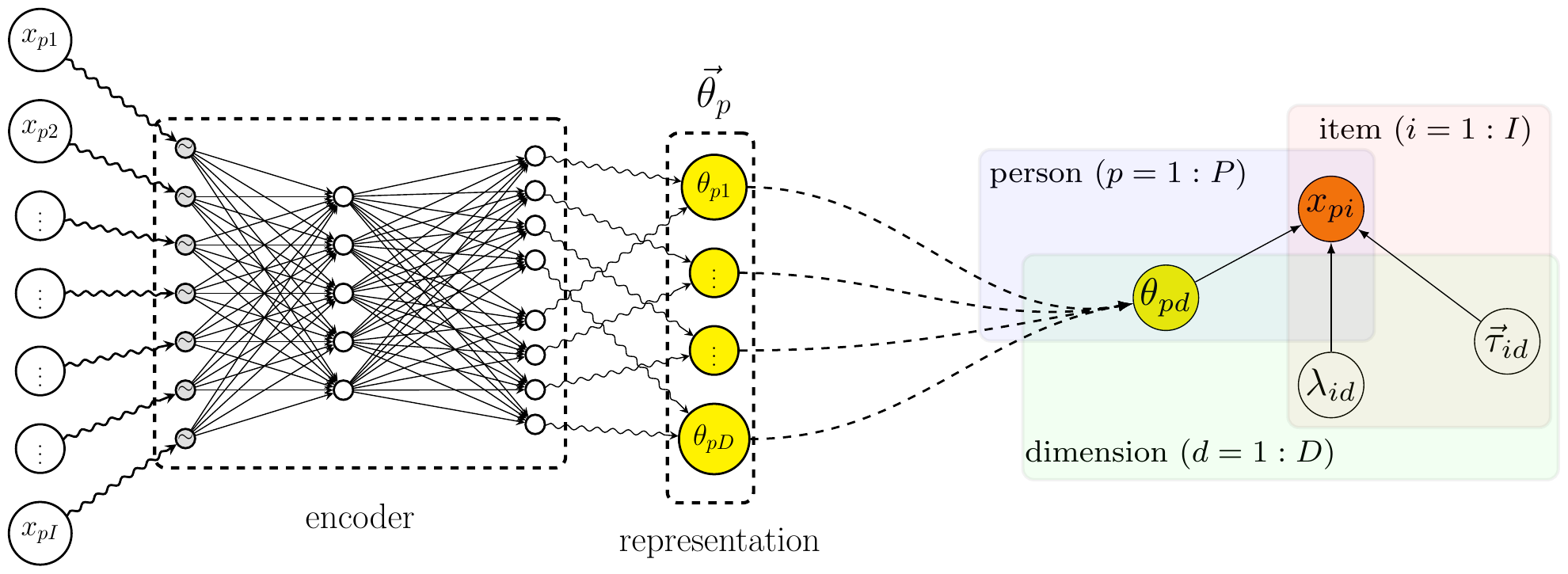}
\caption{\textbf{Architecture:} The encoder (left) computes via an uninterpretable neural network an interpretable representation of the input data consistent with the mechanistic Bayesian hierarchical model (right).}
\label{fig:architecture}
\end{figure}

In the parlance of neural networks, we present the overall method under the framework of a probabilistic autoencoder, depicted in Fig.~\ref{fig:architecture}. Item response theory (IRT) provides a substitute for the less-structured decoder neural network present in vanilla autoencoders.

\subsection{Hierarchical Bayesian item response theory (IRT)}

In item response theory (IRT), a person's test responses are modeled using both personal traits and item-specific parameters.  The item  parameters relate to the difficulty of the item and the discrimination of the item, or the degree to which the question's responses are determined by personal traits. The two types of attributes are linked together via item response functions.

We focus on the parametric graded response model (GRM), which is used to model tests with ordinal responses.
According to the GRM, the probability of a response of $j\in\{1,2,\ldots,J\}$ to item $i$ for person $p$ is the likelihood function
\begin{align}
\Pr( X_{pi} = j \vert \theta_p, \tau_{i,j}, \tau_{i,j+1} , \lambda_i  )  &= \Pr(X_{pi}\geq j \vert\theta_p, \tau_{ij} , \lambda_i )- \Pr(X_{pi}\geq j+1 \vert\theta_p, \tau_{i,j+1} , \lambda_i ) \nonumber\\
&=  S(\lambda_i(\theta_p-\tau_{i,j})) - S(\lambda_i(\theta_p-\tau_{i,j+1})),\label{eq:GRM}
\end{align}
where $S$ is the sigmoid function
\[
S(x)=\frac{1}{1+e^{-x}}.
\]

The item discrimination parameters $\lambda_i$ control the scale of the response functions, or the strengths of the relationships between trait and response: $\lambda_i=0$ implies that a item is non-informative.
The item-specific parameters $\tau_{ij} < \tau_{i,j+1}$  and the personal traits $\theta_p$ set the locations of the response distribution. For notational simplicity, we set $\tau_{i,0}=-\infty$ and $\tau_{i,J} = \infty$.

Extending the GRM to multidimensional traits, we consider a mixture model where each component of the trait is informed by a weighted combination of the original items.
Our objective in setting priors for the training problem is to simultaneously regularize an IRT model and factorize the original test items. To this end, we encourage sparsity in the mappings between the latent representation and the original items, by using hierarchical shrinkage on the item-domain discrimination parameters.

To encourage item-to-domain sparsity, we set horseshoe priors~\cite{piironen_sparsity_2017} on the discrimination parameters, yielding the hierarchical model equations
{
\begin{equation}
    \begin{split}
        &{\textbf{Likelihood} }\\
        \lefteqn{\Pr( X_{pi} = j \vert \{\theta_p^{(d)}\}_d, \{\{\tau_{ij}^{(d)}\}_j\}_d , \{\lambda_i^{(d)}\}_d  )} \\
         &\quad= \sum_{d=1}^D w_{id} \Pr( X_{pi} = j \vert \theta_p^{(d)}, \tau_{i,j}^{(d)}, \tau_{i,j+1}^{(d)}, \lambda_i^{(d)}  ) \\
        &w_{id} = {\left(\lambda_{i}^{(d)}\right)^\nu}/{\sum_{d=0}^{D-1} \left(\lambda_{i}^{(d)}\right)^\nu} \\
        &\vec{\theta}_p\vert \mathcal{W} = \texttt{encoder}(\vec{X}_p ; \mathcal{W}) \\ \\
    \end{split}
    \qquad
    \begin{split}
        \textbf{Prior} \\
        \lambda_i^{(d)} \vert \eta_{i},\xi_{i}^{(d)},\kappa^{(d)}&\sim \mathcal{N}^+\left(0,\eta_{i}\xi_{i}^{(d)}\kappa^{(d)}\right)  \\
        \xi_{i}^{(d)}&\sim C^+(0,\xi_{0}) \\
        \eta_{i}&\sim C^+(0,\eta_{0}) \\
        \kappa^{(d)}&\sim C^+(0,\kappa^{(d)}_0) \\
        \tau_{i,0}^{(d)} & \sim \mathcal{N}(\mu_i^{(d)},1) \\
        \mu_i^{(d)} &\sim \mathcal{N}(0,1)  \\
        \tau_{i,j}^{(d)}|\tau_{i,j-1}^{(d)} &\sim \mathcal{N}^+(\tau_{i,j-1}^{(d)},1)  \\
        \theta_p,\mathcal{W} &\sim \mathcal{N}(\theta_p;0,1)p(\mathcal{W},\ldots),
    \end{split}
    \label{eq:model}
\end{equation}
}
defined for $i\in 1,\ldots,I \quad d\in 1,\ldots, D,\quad p \in 1, \ldots, P$, and $\mathcal{W}$ represents neural-network parameters. 

The half-Cauchy hyperparameters determining the variance of the discrimination parameters constitute a variant of the horseshoe prior.
In this context, the horseshoe prior factors items into domains by encouraging sparsity in the discrimination parameters. We also use horseshoe priors in regularizing the probabilistic encoder.
 When applied to weights of neural networks, the horseshoe prior has been shown to improve out of sample performance~\cite{ghosh_structured_2018,ghosh_model_2017}, as it is also known to do in the general setting of Bayesian hierarchical models~\cite{polson_half-cauchy_2011,polson_shrink_2011}.

\subsection{Probabilistic encoder}

Referring to Fig.~\ref{fig:architecture}, we use the graded response model as a probabilistic decoder. Note that the decoder model is itself sufficient for calibrating and implementing item response theory. However, by coupling a neural network encoder to the decoder, we obtain a test scoring algorithm for a new subject's responses
\begin{equation}
    \texttt{encoder}: \substack {\text{new} \\ \text{responses}}\to \mathbb{E}_{\substack{\text{posterior} \\ \text{parameters}}}\Pr\left( \mathrm{traits} {\bigg\vert} \  \substack{\text{training}\\ \text{responses}},\ \substack{\text{calibrated}\\\text{model}} \right),
    \label{eq:scoring}
\end{equation}
consistent with the Bayesian inference performed in training the IRT model. Note that  Eq.~\ref{eq:scoring} is from the viewpoint that the encoder represents probability densities. There are many ways to accomplish this representation. One way of doing so is to use weight-uncertain Bayesian neural networks~\cite{blundell_weight_2015}, where the encoder is an ensemble of neural networks, in a manner true to Fig.~\ref{fig:architecture}. Other computationally-efficient approximate methods also exist such as dropout~\cite{wen_flipout:_2018,gal_dropout_2015}, or simply using arbitrary neural network architectures to express either parametric~\cite{kingma_auto-encoding_2013}, semiparametric, or non-parametric probability distributions~\cite{kobyz_conditional_2015,likas_probability_2001}.

Looking at implementation details, there are some further similarities between the GRM and neural networks.
The GRM (Eq.~\ref{eq:GRM}) has scaling degeneracies that are absolved by our hierarchical priors. Similarly, neural networks are also scale-degenerate, in the absence of scale-enforcing priors or regularization. Additionally, there is a permutation degeneracy in the GRM, that compromises identifiability, that we ameliorate by setting $\kappa_{0}^{(d+1)}<\kappa_{0}^{(d)}$. The  also exists the same degeneracy in DNNs: each dense layer of $n$ nodes has $n!$ computationally-identical configurations, suggesting a similar remedy.

\subsection{Training}

This model makes use of the horseshoe prior, for which the Cauchy distribution plays a starring role. The heavy peak and fat tails of Cauchy priors give it excellent shrinkage properties, however, they make it difficult to sample from directly. In lieu of direct sampling, we use the following auxiliary parameterization~\cite{makalic_simple_2016},
\begin{equation}
\begin{split}
    x\sim\mathcal{C}^+(0, \sigma)
\end{split}
      \qquad\Longleftrightarrow
 \qquad
      \begin{split}
          x^2\sim \text{Inverse-Gamma}\left(\frac{1}{2},\ \frac{1}{\lambda}\right) \\
          \lambda \sim \text{Inverse-Gamma}\left(\frac{1}{2},\ \frac{1}{\sigma^2}\right).
      \end{split}
\end{equation}

Directly training the conjoined model of Fig.~\ref{fig:architecture} is challenging. Our strategy has been to proceed in blocks, by initially training the IRT model on its own, and then training the neural network separately, conditional on the IRT model. For both sets of training, we used mean-field ADVI with inverse-Gamma distributions parameterizing the scale parameters.

\subsection{Model evaluation}

To decide on a metric for model comparison, we consider the real-world use-case of IRT models. In practice, IRT models are first calibrated to a training set of responses from which both person- and item-specific model parameters are learned. Then, conditional on these pre-trained item-specific parameters, new respondents are scored by learning their person-specific parameters (traits). Hence, the scoring procedure finds the traits of new respondents relative to those of the calibration sample.

Implicitly, the assumption is that the calibrated model is predictive of how new respondents interact with the test. Hence, the out-of-sample predictive power of any calibrated IRT model is important. The general way to assess this power is to use held-out data, through k-fold cross validation, which requires multiple model fits. To save from having to pay the added computational cost of refitting the model, we adapt an approximation of Bayesian leave-one-out cross validation (LOO) known as the Widely Applicable Information Criterion (WAIC)~\cite{watanabe_asymptotic_2010,gelman_understanding_2014,piironen_comparison_2017,vehtari_practical_2017}. 

The WAIC is a simple-to-compute asymptotic approximation of LOO under minimal regularity conditions. In particular, the WAIC does not assume that the mapping between data and model parameters is bijective.
The WAIC consists of the balance between two simple terms which often have exact algebraic expressions~\cite{chang_predictive_nodate}. The first term is the log pointwise predictive density (lppd), which is the ``pointwise'' sum of the logarithm of the marginal likelihoods for the training data, or
\begin{equation}
    \text{lppd} = \sum_p \log\left\{ \mathbb{E}_{\substack{\text{param}\\\text{posterior}}}\left[\Pr\left(\mathbf{X}_p \Big| \text{params} \right)   \right]\right\} \approx \sum_p \log\left\{ \frac{1}{S}\sum_{s=0}^{S-1}\left[\Pr\left(\mathbf{X}_p \Big|\,  \substack{\text{param}\\\text{sample }s} \right) \right\} \right],
    \label{eq:lppd}
\end{equation}
where we treat each person $p$'s collection of responses $\mathbf{X}_p$ as a statistically interchangeable point and assume that the joint posterior distribution of the calibration parameters is represented by a sample of size $S$. The second term in the WAIC is a penalty term of which there are two commonly used variants. The second variant has lower variance than the first~\cite{gelman_understanding_2014}, so we adapt it to our purposes in computing
\begin{equation}
    \text{pwaic} = \sum_{p} \text{Var}_{\substack{\text{param}\\\text{posterior}}}\left[ \log \Pr\left(\mathbf{X}_p \Big|\,  \substack{\text{param}\\\text{sample }s} \right) \right].
    \label{eq:pwaic}
\end{equation}
Eqs~\ref{eq:lppd} and~\ref{eq:pwaic} combine together to constitute the WAIC,
\begin{equation}
    \text{WAIC} = -2\times(\text{lppd} - \text{pwaic}),
\end{equation}
which is on the same deviance scale as used in the AIC.

The WAIC, being a model statistic, is a stochastic quantity that one estimates conditional on the on-hand data. Hence, it should not be treated as  certain and one must also take its estimation error into account. To this end, one may approximate the standard error of the WAIC using the expression
\begin{align}
    \lefteqn{\widehat{\text{se}}\left(\text{WAIC}\right)\approx }\nonumber\\ &\quad\sqrt{P\times\text{Var}_{\text{people}}\left[\log\left\{ \mathbb{E}_{\substack{\text{param}\\\text{posterior}}}\left[\Pr\left(\mathbf{X}_p \Big| \text{params} \right)   \right]\right\} - \text{Var}_{\substack{\text{param}\\\text{posterior}}}\left[ \log \Pr\left(\mathbf{X}_p \Big|\,  \substack{\text{param}\\\text{sample }s} \right) \right] \right] }.
\end{align}

\section{Experiments}

\begin{figure}
\includegraphics[width=1.0\linewidth]{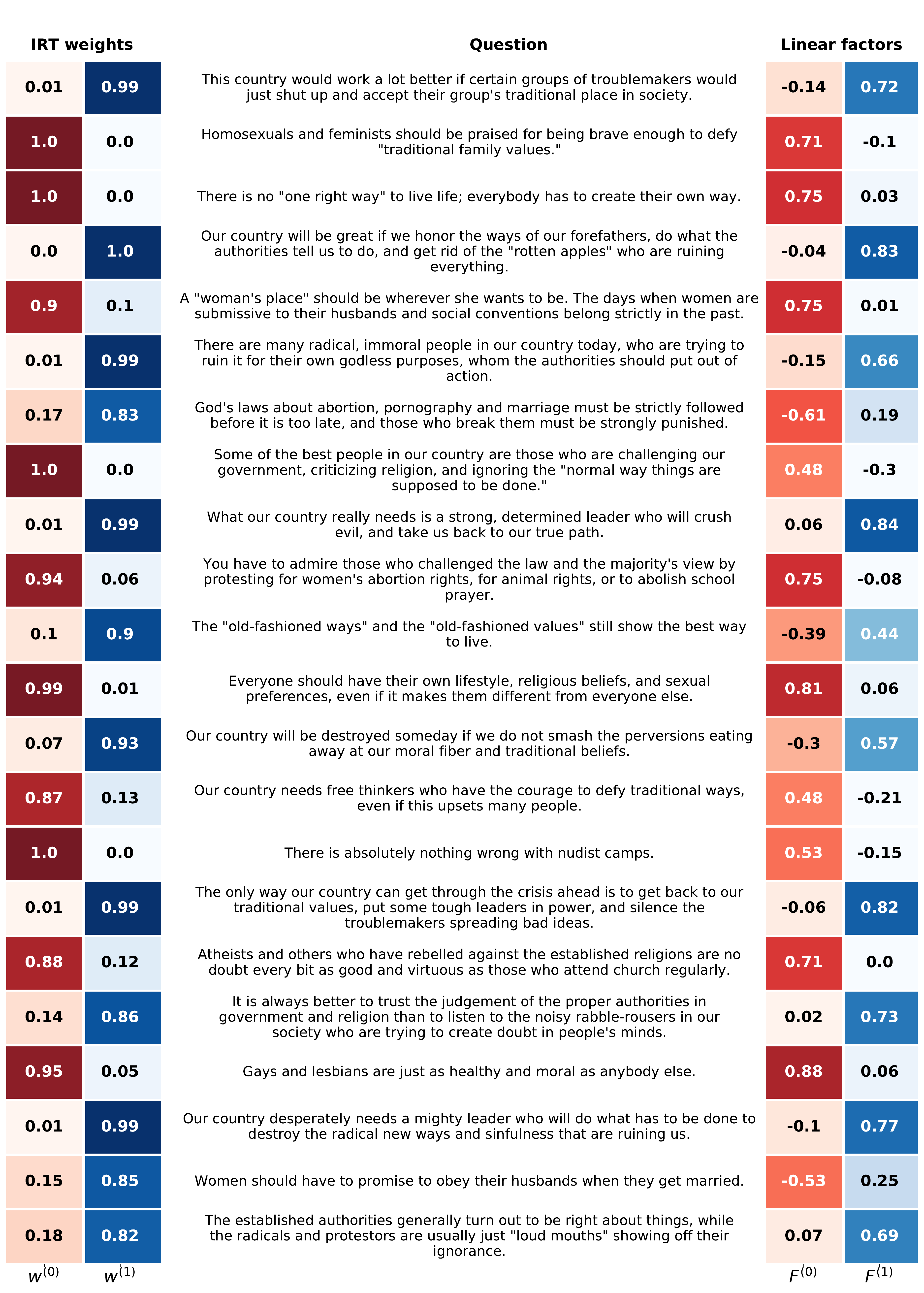}
\caption{\textbf{Two-dimensional within-IRT factorization vs linear factorization of  Right Wing Authoritarianism (RWA) scale test items}, colored by discrimination weights $w_i^{(d)}$. Shown are posterior expectations for the weights as estimated by mean-field ADVI. For comparison, linear factor loadings, colored by absolute value, are contrasted on the right hand side.}\label{fig:RWA2}
\end{figure}

\begin{figure}
\includegraphics[width=1.0\linewidth]{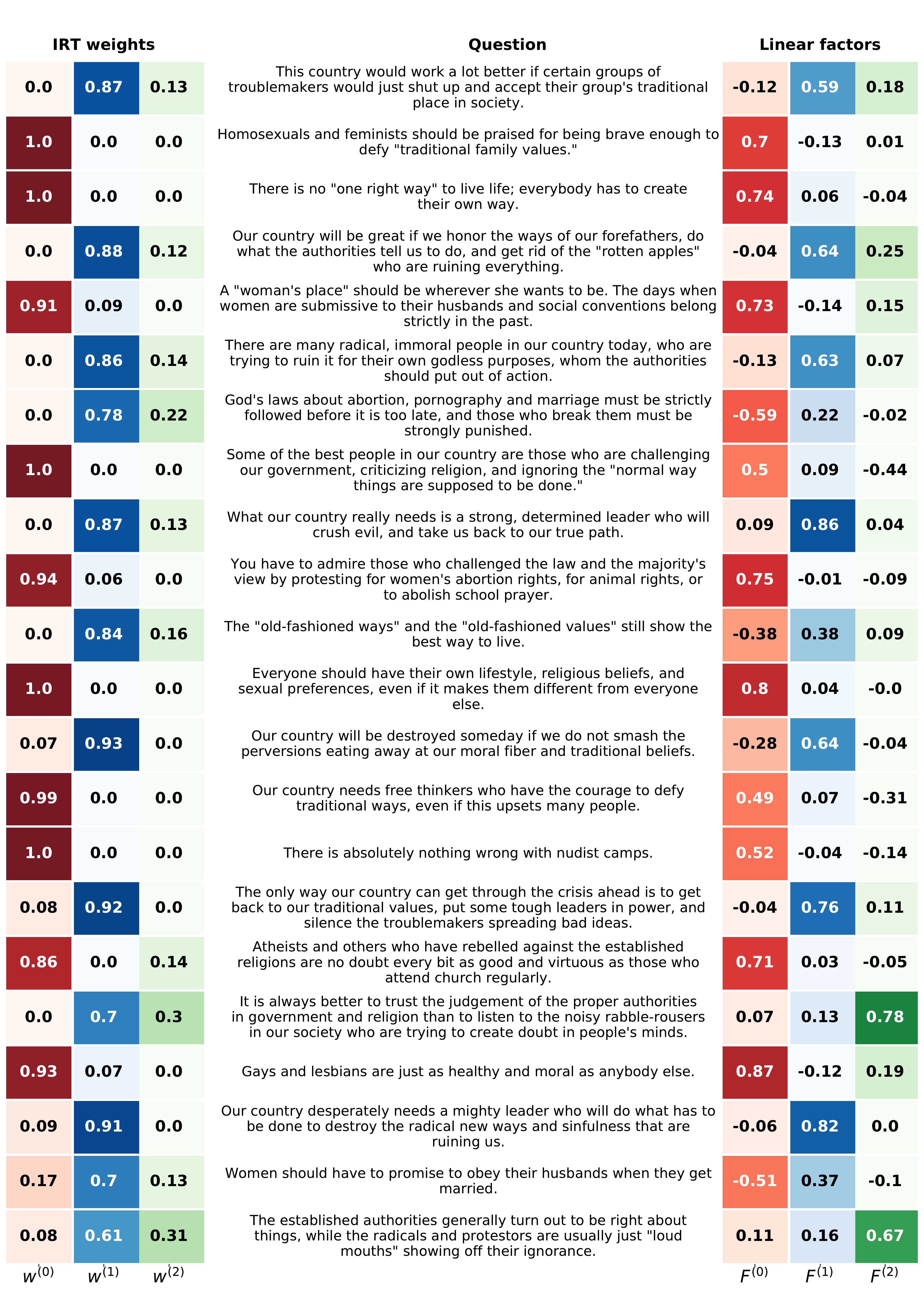}
\caption{\textbf{Three-dimensional within-IRT factorization vs linear factorization of  Right Wing Authoritarianism (RWA) scale test items}, colored by discrimination weights $w_i^{(d)}$. Shown are posterior expectations for the weights as estimated by mean-field ADVI. For comparison, linear factor loadings, colored by absolute value, are contrasted on the right hand side.}\label{fig:RWA3}
\end{figure}


As a testing ground for the overall method, we used the Right Wing Authoritarianism (RWA) scale~\cite{altemeyer_enemies_1988}, for which there exists an openly available dataset~\cite{noauthor_take_nodate} representing $9881$ college-aged respondents. The scale consists of twenty two questions on cultural and societal beliefs, with responses reported on a 9-level Likert scale.
 
Fig.~\ref{fig:RWA2} demonstrates factorization of the scale into two dimensions, learned using mean field ADVI, with inverse-gamma distributions to approximate all scale parameters. In these examples, we set the following scale parameters: $\eta_0=0.01$, $\xi_0=0.01$, $\kappa^{(d)}_{0}=0.01\times (0.1)^{d}$, $\nu=1$.
For comparison, the linear factor loadings are also given. 

Fig.~\ref{fig:RWA3} demonstrates factorization of the scale into three dimensions, also learned using mean field ADVI, with inverse-gamma distributions to approximate all scale parameters. Also in these examples, we set the following scale parameters: $\eta_0=0.01$, $\xi_0=0.01$, $\kappa^{(d)}_{0}=0.01\times (0.1)^{d}$, $\nu=1$.

In both Figs.~\ref{fig:RWA2} and~\ref{fig:RWA3}, we initialized the ADVI algorithm so that the mean of the discrimination parameters is one plus the corresponding factor loading. We found this choice to lead to stable training using the Adam optimizer with an initial step size ranging between $10^{-3}$ and $10^{-4}$, and with an exponentially decaying learning rate.

The WAIC statistics for the two and three-dimensional models were $\text{WAIC}_{D=2} \approx 5.67 \pm 0.025 \times 10^5$ and $\text{WAIC}_{D=3} \approx 5.66 \pm 0.025\times 10^5$ respectively, mildly supporting the adaptation of a three-dimensional factorization on purely predictive grounds.

We performed all experiments using TensorFlow probability~\cite{dillon_tensorflow_2017}. The source code is available upon request. The python package \href{https://github.com/EducationalTestingService/factor_analyzer}{\texttt{factor-analyzer}} provided the algorithm for our exploratory factor analysis results that we contrasted against the within-IRT factorizations.

\section{Discussion}

In this manuscript we have defined a Bayesian hierarchical model for performing factorization of test items in item response theory, making an analogy to probabilistic autoencoders. We demonstrated the method on the Right Wing Authoritarianism scale and used Bayesian predictive model comparison to contrast two and three-dimensional factorizations on the basis of prediction. 


\subsection{Within-IRT versus linear factorization}

In practice, people use factor analysis as a first step in item response theory, in order to segregate items into separate domains for fitting IRT. Factor analysis results in a non-sparse matrix of loadings. To use factor analysis for this purpose, one must adapt an ultimately arbitrary cut-off criteria. A common criterion used in the literature is to set the cutoff at an absolute value of $0.4$, though other advised cutoffs dependent on sample size exist~\cite{hair_multivariate_1998}.

For the RWA factorization, an absolute cut-off of $|F|<0.4$ in the factor analysis leads to a different factorization from that obtained within-IRT, though with some similarities. 
Looking at the two dimensional factorization of Fig.~\ref{fig:RWA2}, 
the within-IRT factorization assigns fewer overall items into the first dimension relative the linear factor analysis. Looking at the item separation within-IRT, there are clear differences between the two overall factor dimensions. The first dimension tends to pick up on items that describe attitudes towards openness to liberal cultural attitudes. The second dimension tends to pick up on items describing adherence to traditional values and deference to authorities.

The two methods clearly disagreed in their categorization of two items: \textit{Women should have to promise to obey their husbands when they get married}, and \textit{God's laws about abortion, pornography and marriage must be strictly followed before it is too late, and those who break them must be strongly punished}. Factor analysis placed both of these items into the first factor, though with negative loading. Within-IRT factorization placed these two items into the second factor instead.

Extending the factorizations to three dimensions (Fig.~\ref{fig:RWA3}), one sees that for the within-IRT factorization, some of the items that weighted nearly completely into the second dimension leak partially into the third dimension, while retaining the bulk of their weight in the second dimension. In contrast, for linear factorization, two of the items switch completely from the second factor to the third factor. This behavior suggests that the IRT factorization is inherently more-stable than linear factor analysis, demonstrating the partial pooling feature of Bayesian hierarchical modeling.

\subsection{Selecting between two and three dimensions}

Based purely on predictive accuracy, the three-dimensional model achieves a slight edge over the two-dimensional one, though the difference between the WAIC statistics for these models was within one standard error.

For practical purposes, the two models have very similar predictive capabilities, as estimated using the WAIC. Hence, other considerations must be used for deciding between the models. These considerations are  highly context-dependent, and should be influenced  by how one wishes to interpret or use the testing instrument.

The interpretability of the two-dimensional model is fairly clear, given the content of the items within the resulting factors.
The two- and three-dimensional models have similar factorizations, except for some cases where the second dimension leaks some weight into the third dimension. The meaningfulness of the third dimension requires subject-matter knowledge to adjudicate.

\subsection{Scoring}

When training a Bayesian model using mean-field variational methods, one is breaking dependencies between the variables of interest. For this reason, the sampling distribution of the surrogate posterior distribution does not necessarily yield the correct prediction distribution, unless either the posterior parameter densities are heavily concentrated around an unimodal peak, or are factorizable. 

The entire purpose of IRT is to obtain statistics of the prediction distribution for latent traits conditional on input responses. Mindful of this purpose, it is improper to directly use the surrogate posterior parameter distributions to perform scoring -- this holds true regardless of whether one wishes to score using the neural network, or by marginalizing over the statistics of the IRT model. In either case, one must use either more-structured variational approximations, or direct Monte-Carlo samplings from the joint posterior density, to preserve at least some of the inter-variable associations. 

Using direct Monte-Carlo simulations of the parameters is conceptually straightforward, yet computationally intensive, both at the time of training and at the time of prediction. One can reduce the computation needed at training by performing MCMC only after ADVI, but then still be left with the computation of propagating the parameter samples at the time of prediction.

Since Bayesian computation is approximate by nature, it is perhaps fruitful to forgo Bayesian neural networks and seek deterministic neural networks that directly parameterize Gaussian or other low-order approximations of the prediction distributions. Doing so would also come with the added benefit of making scoring incredibly fast. Direct Bayesian marginalization of the IRT model requires computation of a multidimensional integral either directly or through Monte Carlo methods. For this reason, the neural network architecture would be advantageous. Future work will focus on finding efficient architectures for this task. 

\section{Conclusion}

We have demonstrated that one may perform multidimensional within-IRT factorization and model calibration in a single step, using sparse coding. By acknowledging the analogy between IRT and probabilistic autoencoders, one also obtains a versatile tool for inverting IRT model products to score new test respondents. On the flip side, regularization used for Bayesian multidimensional IRT shows promise for taming Bayesian neural networks. Future work will concentrate on translating the pooling properties of Bayesian hierarchical models to neural networks.

\section*{Acknowledgements}

This work is supported by the Intramural Research Programs of the National Institutes of Health Clinical Center (CC) and the National Institute of Diabetes and Digestive and Kidney Diseases (NIDDK),  and the US Social Security Administration.

\clearpage
\newpage 

\bibliography{irtvae}

\appendix



\end{document}